\documentclass[runningheads]{llncs}

\usepackage[T1]{fontenc}
\usepackage{hyperref}
\usepackage{graphicx} 
\usepackage[misc]{ifsym}
\usepackage{enumitem}
\usepackage{xcolor}
 \usepackage[most]{tcolorbox}
\usepackage{dblfloatfix}    

\newtcolorbox{definition_}{
enhanced,
boxrule=0pt,frame hidden,
borderline west={2pt}{0pt}{blue!55!black},
colback=blue!5!white,
sharp corners
}

\begin{document}
\title{Model Lake\,: a New Alternative for Machine Learning Models Management and Governance}
\titlerunning{Model Lake\,: a New Alternative for ML Model Management and Governance}
%
\author{Moncef Garouani\inst{1}\textsuperscript{(\Letter)}\and
Franck Ravat\inst{1}\and
Nathalie Valles-Parlangeau\inst{2}}

\authorrunning{M. Garouani et al.}

\institute{  IRIT, UMR 5505 CNRS, Université Toulouse Capitole, Toulouse, France\\
\email{moncef.garouani@irit.fr, franck.ravat@irit.fr}\and
 LIUPPA, Université de Pau et des Pays de l’Adour, Anglet, France\\
\email{nathalie.valles-parlangeau@iutbayonne.univ-pau.fr}
}
\maketitle              
\begin{abstract}
The rise of artificial intelligence and data science across industries 
underscores the pressing need for effective management and governance of machine learning\,(ML) models. 
Traditional approaches to ML models management often involve disparate storage systems and lack standardized methodologies for versioning, audit, and re-use. Inspired by data lake concepts, this paper develops the concept of ML \textit{Model Lake} as a centralized 
management framework for datasets, codes, and models within organizations environments. We provide an in-depth exploration of the Model Lake concept, delineating its architectural foundations, key components, operational benefits, and practical challenges. We discuss the transformative potential of adopting a Model Lake approach, such as enhanced model lifecycle management, discovery, audit, and  reusability. Furthermore, 
we illustrate a real-world application of Model Lake and its transformative impact on data, code and model management practices.

\end{abstract}

\section{Introduction}

The rapid advancement of artificial intelligence\,(AI) technologies and transformative adoption of data science have emerged as a force across diverse domains\,\cite{jamt}. From predictive analytics to natural language processing, AI technologies are driving innovation and 
unlocking new opportunities for organizations worldwide to gain insights, automate processes, and deliver personalized experiences\,\cite{Kreuzberger_2023}. However, as the adoption and 
volume of AI applications continues to grow, so does the complexity associated with effectively managing and leveraging the power of the developed ML models\,\cite{pal2024}.

Traditionally, managing and governing ML models within enterprise environments has been a fragmented and labor-intensive process\,\cite{Amershi_2019}. Data scientists and engineers often encounter challenges related to lifecycle management, governance, and re-use of models across different systems and environments\,\cite{Kreuzberger_2023}. Furthermore, as the number of models proliferates within an organization, the task of  tracking, and maintaining these models becomes increasingly challenging\,\cite{Traub2021}. Traditional approaches often involve multiple storage systems, fragmented workflows, and a lack of standardized methodologies for versioning and monitoring trained models and mined datasets. This fragmented landscape poses significant obstacles to organizations to effectively harness the full potential of their model repositories\,\cite{Kreuzberger_2023}. Consequently, there is a pressing need for a more unified and efficient approach to model management within organizations environments.

In response to these challenges and inspired by the success of collaborative data sharing platforms, such as Data Hub\,\cite{Bhardwaj_2015}, and the development of data lakes\,\cite{Ravat2019} in storing and managing vast volumes of heterogeneous data that can feed analytics, AI or data science services, a new paradigm known as the ``\textit{Model Lake}" has emerged as a potential solution to the challenges of ML model management\,\cite{pal2024}. Similar to a data lake, Model Lake aims to provide a unified platform for storing, organizing, and governing the lifecycle of ML models. The concept of a Model Lake, in our definition, goes beyond mere storage and encompasses a suite of functionalities designed to streamline the entire lifecycle of data analysis pipeline, from data ingestion and preparation to model development and deployment. By centralizing model management activities within a single ecosystem, organizations can improve model reuse, foster collaboration among data scientists and engineers, and accelerate the deployment of AI solutions\,\cite{Kreuzberger_2023}.

The concept of the Model Lake has recently emerged, despite previous proposals for Model registry solutions by IT companies and limited academic envisions\,\cite{Traub2021,pal2024}. Nevertheless, there lacks a standardized definition or recognized architecture for a model lake. In this work, we explore this emerging concept for addressing the issue of managing and maintaining productive ML models. We envision this ecosystem playing a dual role as both a centralized hub and a dynamic workspace for ML model management and governance. As a centralized hub, the Model Lake serves as a registry for storing, versioning, and governing ML models, akin to a ``Github" for models where models are cataloged, annotated, and easily accessible to stakeholders across the organization. This centralized approach streamlines model discovery and access, reduces duplication of efforts, and ensures consistency and reproducibility across the model lifecycle. Simultaneously, as a dynamic workspace, the Model Lake provides a collaborative environment where data engineers and data scientists can seamlessly collaborate, experiment, and iterate on mined datasets and developed models in real time. This dual functionality not only promotes agility but also enables better reuse the models, monitor the performance of deployed models, and track changes in data and models characteristics. By bridging the gap between data engineering, model development and serving, the Model Lake empowers organizations operational efficiency and governance of their entire data analysis pipelines.

\section{Background and Related Work}

Artificial Intelligence is transforming various industries with an increasing number of ML models being utilized to solve complex real-world problems\,\cite{Garouani2021,Chaabi_iaes}. Every day, thousands of new ML models are conceptualized, trained, and deployed, each with its own distinct design, training data, and functionalities\,\cite{Garouani_mtl}. This expansion in the variety and in the efficiency of models leads to remarkable advancements, but also brings about a range of pressing challenges\,:

\begin{itemize}
    \item \textit{How do we effectively find, understand, and manage this heterogeneous collection of many available ML models?}
    \item \textit{How do we understand what a model does, how it was trained, and how it relates to other models?}
    \item \textit{How can we ensure the ethical development and application of developed models in situations where mined data and learned models lineage is opaque?}
\end{itemize}

With the advance of open source libraries, such as scikit-learn\,\footnote{\url{https://scikit-learn.org}}, Tensorflow\,\footnote{\url{https://www.tensorflow.org}}, and PyTorch\,\footnote{\url{https://pytorch.org}}, training a ML model has become much more approachable\,\cite{garouani_IDEAL23}. 
Nevertheless, in practice, getting a model trained is only the start of the journey. 
The ML life-cycle has different methodologies to fit different scenarios and data types. It involves manual steps for deploying the ML pipeline model. This method can produce unexpected results due to the dependency on data, preprocessing, model training, validation, and testing.  Consequently, managing and governing this process poses a challenge for IT companies\,\cite{Amershi_2019,Schelter2018OnCI}. To address the management problems, they have developed systems known as the Model Registry streamlining the process of deploying ML models into production. These systems serve as centralized repositories where teams can share their ML models. The model registry concept imitates traditional software package registries such as Github. Through web searches, we identified several model registries, including Hugging Face\,\cite{huggingface}, TensorFlow Hub\,\cite{tensorflowHub}, and ONNX Model Zoo\,\cite{onnxONNXModel}. Among all registries, Hugging Face offers the largest and most diverse set of pre-trained ML models, hosting over 600\,000 model.

While ML Model Registries address certain aspects of model management, they also come with limitations. Model registries typically adopt a schema-on-read approach, where models and their properties are shared. Yet, without implicit data management and governance, the ingestion of diverse models—each tailored to specific tasks, data, and contexts can lead to a ``model swamp" rendering it opaque, inaccessible, and unreliable to users. One limitation is that Model Registries primarily focus on storing individual models, which may lead to fragmentation and soloed repositories. 
W. Jiang et al.\,\cite{jiang2023} investigated model reuse in the Hugging Face Model Registry. Initially, they conducted interviews with 12 Hugging Face practitioners to understand the practices and challenges of model reuse. Additionally, they analyzed how 63182 models are created and shared within the registry. Their findings highlighted several challenges related to model reuse, such as missing attributes, disparities between claimed and actual performance metrics, as well as risks such as privacy concerns and ethical dilemmas arising from the lack of transparency in mined data lineage. 

More recently, W.Liang et al.\,\cite{liang2024whats} conducted a comprehensive analysis of 74\,970 ML model cards\,\footnote{A common semi-structured form of model documentation.} uploaded by 20\,455 distinct user accounts on the Hugging Face model registry. Their findings indicate that only 32\,111 (44.2\%, contributed by 6392 distinct user accounts) out of the 74\,970 model repositories currently include model cards as unstructured Markdown README.md files within their model repositories, making more than 56\% of models among the studied collection unreliable. This substantial number of useless models underscores the need for a more data-centric approach that can enhance the quality and reliability of models within model registries and foster advancements in responsible AI research and development.

As organizations continue to expand their ML initiatives and embrace emerging technologies such as deep learning and reinforcement learning, there arises a need for a holistic and scalable management infrastructure that can adapt to evolving requirements. This is where the concept of Model Lake becomes essential\,\cite{pal2024}. Similar to a data lake, a Model Lake would serve as a centralized ecosystem for storing, organizing, and governing all elements related to ML pipeline, including data, models, code, metadata, and experimental results.

\section{Model Lake}

\subsection{Model Lake Definition}

The Model Lake concept, though emerging recently, lacks a standardized definition or recognized architecture despite existing proposals from IT companies for what a Model Lake should look like, often referred to as Model Registry solutions. These proposals likely vary in terms of features, architecture, and integration capabilities, reflecting the diversity of needs and contexts in which such a system might be deployed\,\cite{Kreuzberger_2023}. To the best of our knowledge, only two academic \textit{vision works} by J. Traub, et al.\,\cite{Traub2021} and K. Pal et al.,\cite{pal2024} have introduced the initial concept. They envision a unified asset ecosystem that brings together data-related assets, including data, algorithms, models, and computational resources and provides them to a broad audience. K. Pal et al.,\cite{pal2024} define model lakes as robust repositories for managing heterogeneous models and their associated meta-data. Their envision seeks to expand and integrate efforts on model provenance, citation, and version management within a centralized ecosystem. However this \textit{model-centred} vision overlooks the life-cycle of the mined data, which is crucial for assessing the reliability and, consequently, the reusability of the stored models.

To be as complete as possible, we extend the definitions of data lakes as proposed by (F. Ravat et al., 2021)\,\cite{Ravat2019} and model lakes as envisioned by (K. Pal et al., 2024)\,\cite{pal2024}. We propose an integrated ecosystem designed to facilitate the ingestion, development, management and the governance of big data analytics and machine learning models within organizations environments.

\begin{definition_}
\textbf{Definition.} Model Lake stands as an integrated ecosystem encompassing respectively the input, process, output and governance aspects of both mined data and developed models. It acts as a centralized hub and management system accommodating diverse data and model types, meeting the requirements of various stakeholders including data engineers, data scientists, data analysts, and business intelligence professionals.
\end{definition_}

In line with our conceptualization, a model lake should offer functionalities for raw data ingestion, on-demand data processing, storage of processed data, data governance, model training and fine-tuning, alongside review, monitoring, and governance of models, associated metadata and code. To address the limits observed in current model registries, we advocate for a generic functional model lake architecture. These functionalities not only ensure data and model provenance but also facilitate data and model management and governance.

\subsection{Model Lake Architecture}

\begin{figure*}[btp] 
    \centering
    \includegraphics[width=1\textwidth]{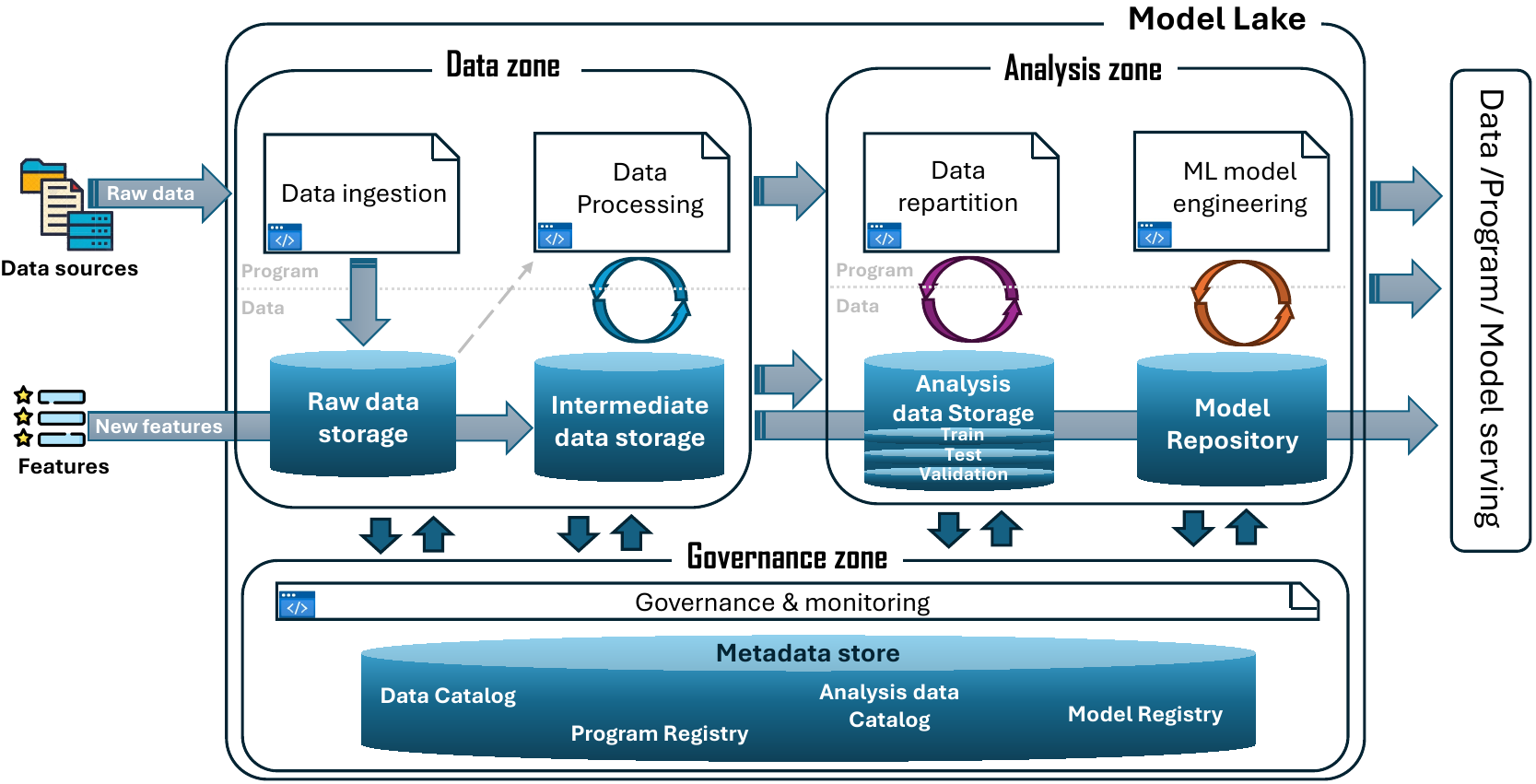} 
    \caption{The proposed Model Lake architecture.}
    \label{fig:archi}
\end{figure*}

To date, there has been a gap in academic exploration concerning the technical architectures of model lakes. Initially, the concept of model repositories, also known as model registries, proposed a functional architecture characterized by a flat structure with a single zone, as suggested by IT organizations\,\cite{Kreuzberger_2023}. This zone facilitates the storage and versioning of trained ML models, covering critical aspects such as model lineage, versioning, tagging, and annotations. However, this architecture falls short in accommodating the storage of the mined data lifecycle, thereby complicating the retrieval of data sources and associated properties. Moreover, the lack of data governance within the model repository undermines data security and quality assurance, rendering model audits unfeasible. In this paper, and in line with our model lake definition, we propose the functional architecture, depicted in Figure\,\ref{fig:archi}, designed to address the limitations of the model repository by offering enhanced capabilities for managing both models and data within the lake ecosystem.

The proposed Model Lake functional architecture comprises three essential zones, each consisting of two layers\,: the processing layer\,(program) and the storage layer\,(data) along with their meta-data.

\begin{itemize}
    \item \textbf{Data Zone}\,: This zone combines both the ingestion and processing stages. Here, all types of data are ingested either in their native formats without immediate processing or as extension of previously processed raw data\,(e.g., to add new features). Big Data ingestion involves connecting to various data sources, extracting data and features, and tracking changes. Subsequently, users can transform raw data into standardized formats according to their requirements. Data processing is an essential step for data analysis and is composed of a set of operations such as data integration, cleaning, transformation, and reduction. This zone involves intermediate processed data storage along with related metadata to ensure lineage tracking.

    \item \textbf{Analysis zone}\,: This zone is a central environment for comprehensive data exploration and ML model development. It enables users to (i)\,perform advanced data exploration\,(meta-features of the dataset, descriptive information about the attributes and the performed transformations), (ii)\,develop and evaluate ML models using various algorithms and hyperparameters tuning techniques. Additionally, it facilitates data repartitioning for training, validation, and testing purposes, as well as model deployment into production environments. Continuous monitoring and feedback loops ensure the ongoing performance and reliability of deployed models. Additionally, programs and metadata storage ensure models lineage tracking, comparison, diffing, as well as auditing and compliance. 

    \item \textbf{Governance and Management Zone}\: This zone is responsible for ensuring Data, Program, Model security, lifecycle management, access, and metadata management. It is crucial to prevent a Model Lake from deteriorating into a model swamp, which is a massive data repository inaccessible to end-users. For all data analysis pipelines, the "Metadata Store" records metadata for each orchestrated ML workflow task. Metadata storage is required for each job iteration on data, programs, and models (e.g., training date, and sources of artifacts). Additionally, model-specific metadata called "model lineage", combining the lineage of data, model, and code, is tracked for each newly registered model. This includes data and model-specific metadata (e.g., source and version of feature data, used parameters, resulting performance metrics), ensuring the full traceability of runs.
\end{itemize}

By adopting the model lake management and governance concept, users can track the complete provenance of a model, the lifecycle of datasets utilized for training, and the specific adjustments or fine-tuning steps that led to its current configuration. This provenance information is invaluable for\,:
\begin{itemize}

\item \textbf{Auditing and Compliance}\,: Ensuring that a model was trained and adapted using approved datasets and methodologies, in adherence to relevant regulations and policies\,(e.g., RGPD\,\cite{RGPD}).

\item \textbf{Reproducibility}\,: Replicating the precise procedures and data employed in model engineering, facilitating reproducible development, search and re-use.

\item \textbf{Bias and Fairness Analysis}\,: Examining the lineage to pinpoint potential sources of bias introduced through the training data or training process.

\item \textbf{Data and Model Evolution}\,: Understanding how data and models have evolved over time, and comparing the models performance or behavior across different data/configurations versions.
\end{itemize}

\subsection{Model Lake Metadata Management}

Similar to data lakes, a model lake is characterized by a "schema on read" principle, allowing the ingestion of diverse datasets and models without a predefined schema. This flexibility enables various transformations and analyses, by different users (data scientists, engineers, analysts). However, this versatility poses a risk of turning the model lake into a model swamp, where data and models become invisible, inaccessible, and incomprehensible. To mitigate this, model lake governance, particularly metadata management, is essential. Metadata 
provide descriptions of resources within the model lake. Information about each step of the ML pipeline is recorded in order to help with artifacts lineage, reproducibility, and comparisons.  A unified schema for metadata, transparently shared among users, and a robust system for managing metadata of ingested data, applied programs, and learned models are essential for effective governance.



\subsubsection{Metadata Model on Data zone}

To ensure the accessibility and reusability of ingested and processed data across various user groups, a comprehensive metadata storage and management system within the "data zone" is imperative. This metadata should contain detailed information at different data analysis stages. In our approach, we adopt the 5W1H\,(What, Who,
Where, When, why, how) method to facilitate a systematic understanding of data ingestion and processing. This method prompts the following inquiries:

\begin{itemize}
\item \textbf{What}: Identifying external data sources and the nature of ingestion activities 
(ingested datasets, their quality, security level, and interrelations).
\item \textbf{Who}: Determining ownership of the source data, as well as the individuals responsible for data ingestion and processing.
\item \textbf{Where}: Locating the storage sites for ingested and processed datasets and associated data ingestion/processing code.
\item \textbf{When}: Establishing timelines for the ingestion and processing of datasets.
\item \textbf{Why}: Understanding the purpose behind the data processing activities.
\item \textbf{How}: Understanding the ingestion and processing operations.
\end{itemize}

\begin{figure}[h!] 
    \includegraphics[width=1\textwidth]{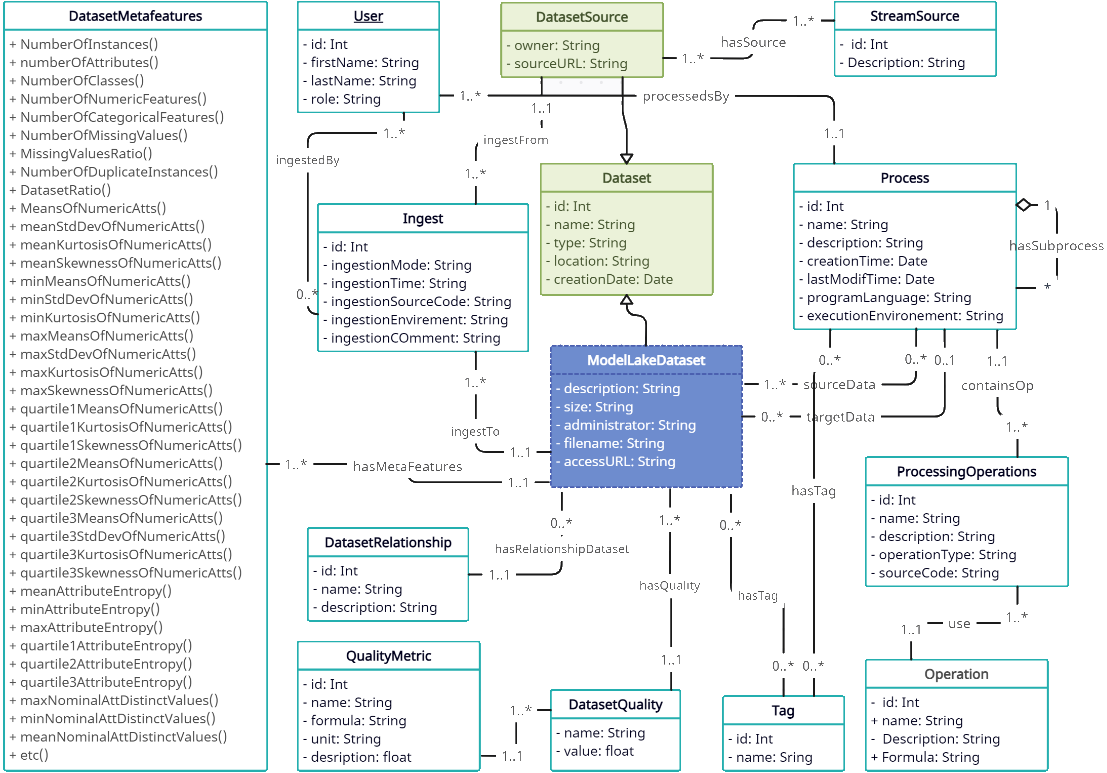} 
    \caption{Metadata Model on Data Ingestion.}
    \label{fig:Metadata_ingestion}
\end{figure}

These questions span four main categories\,: general knowledge about the data catalog, external data sources, data ingestion/ processing activities, and ingested/processed datasets. 
To address these questions and consider diverse data sources and processing methods, we propose the metadata model on data ingestion and processing shown in Figure\,\ref{fig:Metadata_ingestion}.

The introduction of external data sources metadata within the \textit {DatasetSource} class (a subclass of \textit {Dataset}) addresses inquiries about data origins, aiding users in verifying the source of ingested data. This metadata, highlighted in green, includes the name and type of the dataset, a description of the data source (addressing "what"), owner information (answering "who"), location details (answering "where"), and creation date (answering "when"). For understanding data ingestion activities and facilitating the reuse of ingestion processes, we provide metadata concerning the ingestion process in the \textit {Ingest} class and its associated relationships. This metadata includes ingestion mode, comments, relationships (ingest-From, ingest-To), the "ingestedBy" relationship linked to the \textit{User} class (answering "who"), access URL (answering "where"), ingestion time (answering "when"), and ingestion mode and environment (answering "how").

The ingested dataset characteristics metadata aim to help users easily discover, access, and understand datasets without direct inspection. This is achieved through classes such as \textit{ModellakeDataset, DatasetMetafeatures}, and \textit{Tag}, which include attributes like name, format, description, tags, linked entity class, attributes, location, and creation date. Technical process metadata, modeled through attributes in classes such as \textit{Ingest, Process} and \textit{User}, include name, program language, creation time, last modification time, and relationships like source data, target data, and real-time processes. This aids data wranglers in understanding "how, where, when", and by "whom" data are ingested or processed. Attributes like program language, creation time, and source data relationships facilitate understanding and access to processing program codes, enabling reuse or modification. In addition to technical details, Business metadata, stored in classes like \textit {Process}, include descriptions aimed at elucidating the business objectives of performing particular processes (answering "why"). Processing metadata, modeled through the \textit{Process} and \textit{ProcessingOperations} classes, describe the details of processing operations, aiding data wranglers in quickly grasping the content of a process without needing to examine the program's source code in detail.



\subsubsection{Metadata Model on Data Analysis}



\begin{figure}[!h] 
    \includegraphics[width=1\textwidth]{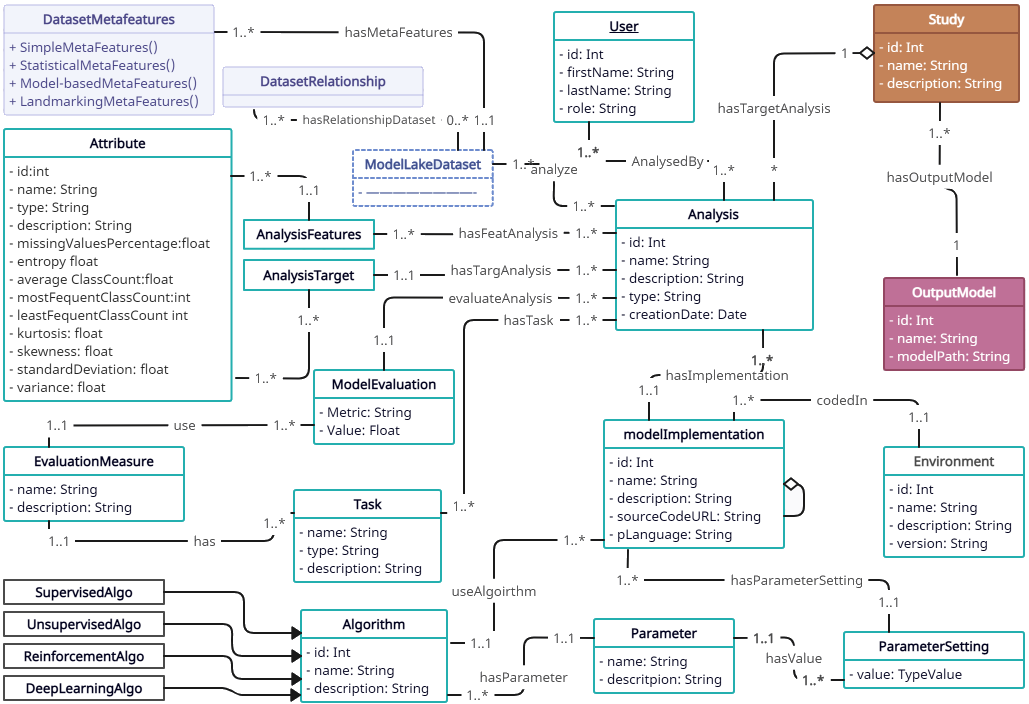} 
    \caption{Metadata Model on Data Analysis.}
    \label{fig:Metadata_analysis}
\end{figure}

To ensure the accessibility of analyses and models, metadata for data analysis should include comprehensive details about users activities. Using the 5W1H method, we address the key aspects needed to search for, understand, and reuse analysis. Analytical metadata enhances collaboration and reuse in ML studies by providing insights not only on datasets but also on previous analysis (Figure\,\ref{fig:Metadata_analysis}). This enables the reuse of existing studies or improvement of ongoing ones, covering key attributes such as implementation details, used features, target classes, models, and performance.

In order to facilitate the reusability within the data analysis pipeline, we introduce analysis metadata to streamline analysis, data, and model discovery. Modeled within \textit{Analysis}, \textit{Study}, \textit{Task}, \textit{Tag}, and \textit{User} classes, these metadata offer vital insights for organizing and comprehending previous analytical projects. A study represents a project consisting of analyses sharing the same subject, while an analysis may be linked to a task addressing specific project objectives. Key attributes, such as the \textit{description} of analysis, studies, and tasks, along with their \textit{type}, provide semantic information about the analysis goals, offering clarity on the "why" behind the work. Additionally, the \textit{User} associations identify "who" conducted the analysis.
The attribute \textit{modelPath} indicates the location of the model and its corresponding implementation source code, addressing the question of "where" the resources are stored. Attributes such as \textit{languageProgram} and information from the \textit{Environment} class elucidate the implementation process. Additionally, details from the \textit{Algorithm} and \textit{Parameter} classes provide insights into the algorithms employed, addressing the question "how" the analysis was conducted. To help users evaluate the reliability of models, we propose the dataset relationship metadata established in the preceding meta-model.

\section{Model Lake Management System}

\begin{figure}[t] 
    \includegraphics[width=1\textwidth]{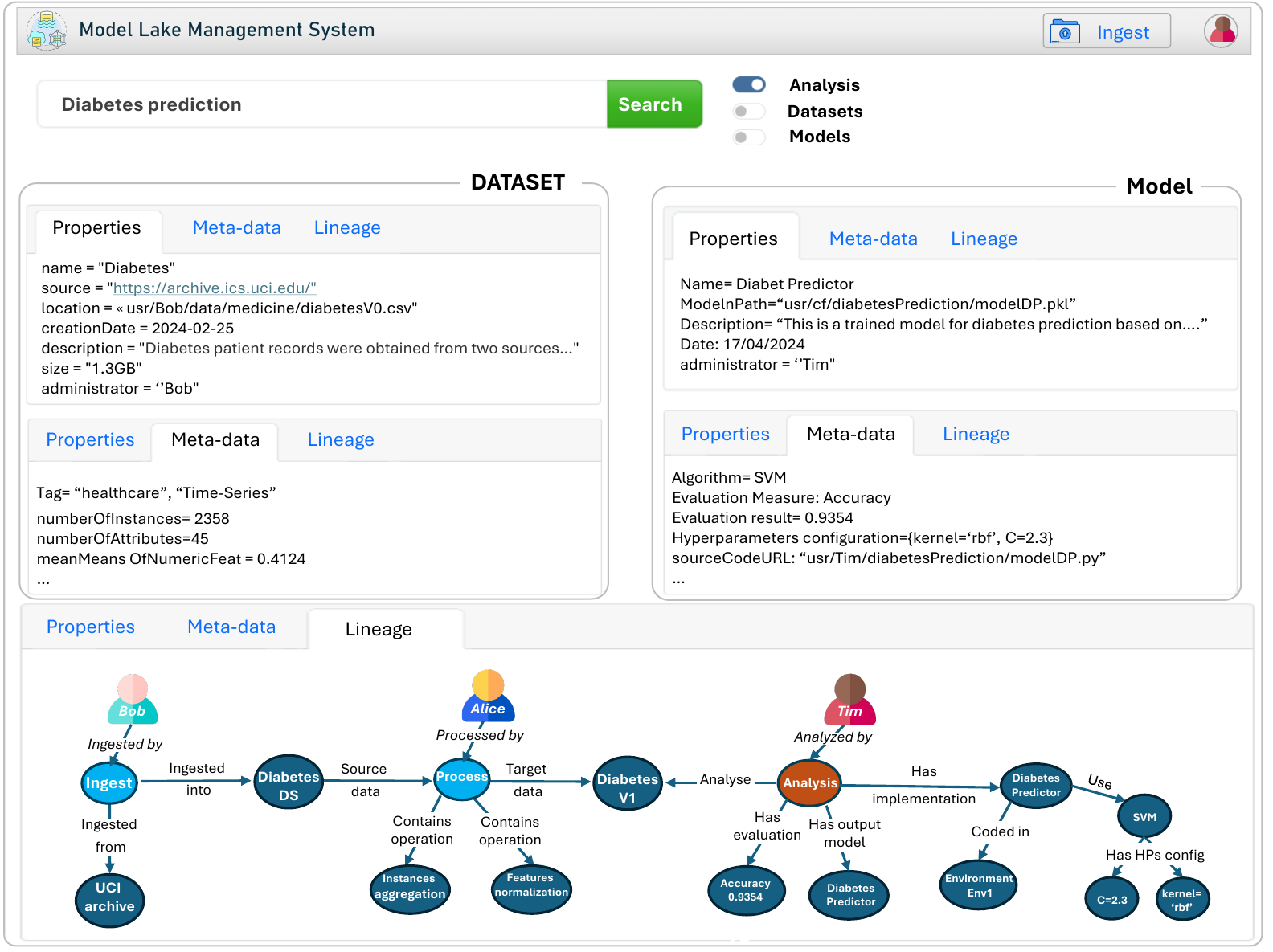} 
    \caption{Model lake management system.}
    \label{fig:interface}

\end{figure}

While the concept of model lake is remaining powerful, realizing its full potential requires careful architectural design and implementation.
At the core of a model lake is the need to store and manage large volumes of datasets, models, code files and associated metadata. This requires a scalable and distributed storage solution that can handle the distinct characteristics of the various ML pipeline artefacts, such as large file sizes, frequent versioning, and complex metadata structures. 
For storing actual model files, including weights and architectures, distributed file systems like the Hadoop Distributed File System or distributed databases are viable options. As models are developed, trained, and updated, they must be ingested into the model lake consistently and in a version-controlled manner. This requires robust ingestion pipelines and versioning mechanisms to ensure that both model files and their metadata are accurately captured and tracked over time. Proper artifact versioning is essential for tracking changes, enabling rollbacks, and maintaining provenance information. Different approaches, such as file-based versioning and metadata-driven versioning, can be employed to manage version information and relationships between artifact versions, effectively creating a versioned knowledge graph (figure\,\ref{fig:interface}- Lineage view).
While artifact files can be stored in distributed file systems or object stores, managing their associated metadata 
requires a separate metadata management solution. Appropriate  solutions include graph databases or distributed key-value stores.

In summary, building an effective model lake metadata management system with an intuitive interface enables users to ingest and retrieve all the ML pipeline artifacts seamlessly. The user interface, as shown in Figure\,\ref{fig:interface}, illustrates how a user can search for previous analysis within the model lake. This interface is designed for a retrieving result of a project titled "Diabetes prediction" and is divided into three main sections\,: Dataset, Model, and Lineage, each providing specific properties, metadata, and lineage information.

\section{Conclusion}
The rapid proliferation of ML models across industries is both an opportunity and a challenge. While the growing diversity and sophistication of models are driving incredible innovations, the lack of standardized management and governance practices risks limiting the full potential of those models. Model lakes offer a promising solution to this challenge, providing a centralized and structured approach to storing, discovering, and managing ML models at scale. However, realizing the full potential of model lakes requires more than just technology. It requires a collaborative effort across the data analysis ecosystem, from researchers and tool developers to data scientists and governance teams. It requires a willingness to adopt new ways of working and a commitment to responsible and transparent AI practices. 
For forthcoming work, we intend to continue the implementation of the model lake management system. This involves expanding its scope to encompass additional types of analysis and ML pipeline artifacts. Additionally, we intend to develop a recommender system to improve data and model search and discovery.

%
%
%
 \bibliographystyle{splncs03_unsrt}
\bibliography{bibliography}

\end{document}